\documentclass{article}
\usepackage{arxiv}

\usepackage[utf8]{inputenc} 
\usepackage[T1]{fontenc}
\usepackage{hyperref}     
\usepackage{url}       
\usepackage{booktabs}       
\usepackage{amsfonts}     
\usepackage{nicefrac}      
\usepackage{microtype}      
\usepackage{url}
\usepackage{makecell}
\usepackage[table, dvipsnames]{xcolor}

\usepackage{tikz}
\usetikzlibrary{
    positioning,      
    fit,              
    decorations.pathreplacing, 
    arrows.meta,       
    calc,shapes.geometric, shadows, backgrounds
}

\usepackage{wrapfig}

\usepackage{dsfont}
\usepackage{amsmath}
\usepackage{amsthm}
\usepackage{amssymb}
\usepackage[bb=boondox]{mathalfa}

\usepackage{algorithm}
\usepackage{algorithmic}
\usepackage{tabularx}

\newcommand{\x}{\mathbf{x}}

\usepackage{graphicx}
\usepackage{subcaption}
\usepackage{tcolorbox}
\usepackage{helvet}
\usepackage{multirow}
\usepackage{titletoc}
\usepackage{mathtools}
\usepackage{enumitem}

\usepackage{hhline}
\allowdisplaybreaks[4]

\newcommand{\figfont}{\fontfamily{bch}\selectfont}

\newtcolorbox{mybox}{
    colback=blue!5, 
    top=1pt,        
    bottom=1pt, 
    left=1.5pt, 
    right=1.5pt
}

\usepackage{lipsum}		
\usepackage{doi}

\def\tran{^{\mathsf{T}}}

\def\one{\mathds{1}}

\newcommand{\bp}{ \begin{proof}}
	\newcommand{\ep}{\end{proof} }

\newcommand{\be}{\begin{equation}}
\newcommand{\ee}{\end{equation}}
\newcommand{\bqq}{\begin{eqnarray}}
\newcommand{\eqq}{\end{eqnarray}}
\newcommand{\bal}{\begin{align}}
\newcommand{\eal}{\end{align}}
\newcommand{\bqn}{\begin{eqnarray*}}
	\newcommand{\eqn}{\end{eqnarray*}}
\newcommand{\nn}{\nonumber}
\newcommand{\ba}{\left[ \begin{array}}
	\newcommand{\ea}{\\ \end{array} \right]}

\def\i{{\boldsymbol{i}}}
\def\j{{\boldsymbol{j}}}

\def\x{{\boldsymbol{x}}}
\def\y{{\boldsymbol{y}}}

\newcommand{\cA}{{\mathcal{A}}}

\newcommand{\cK}{{\mathcal{K}}}

\newcommand{\cO}{{\mathcal{O}}}

\newcommand{\beqn}{\begin{eqnarray}}
\newcommand{\eeqn}{\end{eqnarray}}

\DeclareFontFamily{U}{mathx}{\hyphenchar\font45}
\DeclareFontShape{U}{mathx}{m}{n}{
	<5> <6> <7> <8> <9> <10>
	<10.95> <12> <14.4> <17.28> <20.74> <24.88>
	mathx10
}{}

\def\real{{\mathbb{R}}}

\def\Zint{{\mathchoice{\setbox1=\hbox{\sf Z}\copy1\kern-.75\wd1\box1}
		{\setbox1=\hbox{\sf Z}\copy1\kern-.75\wd1\box1}
		{\setbox1=\hbox{\scriptsize\sf Z}\copy1\kern-.75\wd1\box1}
		{\setbox1=\hbox{\scriptsize\sf Z}\copy1\kern-.75\wd1\box1}}}

\newcommand{\xo}{{\color{blue} \mathbf{x}_o}}
\newcommand{\yo}{{\color{blue} \mathbf{y}_o}}

\newcommand{\xu}{{\color{orange} \mathbf{x}_u}}

\newcommand{\hu}{{\color{orange} \mathbf{h}_u}}

\usepackage{natbib}

\title{Why Ghost Outputs Teach: A Kernel-Based Understanding of Subliminal Learning}

\date{} 			

\author{Zhe Li$^{\dagger}$, \;Bicheng Ying$^{\ddagger}$, \;Chaosheng Dong$^{\mathsection}$, \;Haibo Yang$^{\dagger}$ \\
	$^{\dagger}$Rochester Institute of Technology, \;$^{\ddagger}$ Google Inc., \;$^{\mathsection}$Independent Researcher\\
	\texttt{\{zl4063, hbycis\}@rit.edu, ybc@google.com, ensteindcs@gmail.com} \\
}

\renewcommand{\shorttitle}{Why Ghost Outputs Teach: A Kernel-Based Understanding of Subliminal Learning}

\hypersetup{
pdftitle={A template for the arxiv style},
pdfsubject={q-bio.NC, q-bio.QM},
pdfauthor={David S.~Hippocampus, Elias D.~Striatum},
pdfkeywords={First keyword, Second keyword, More},
}

\begin{document}
\maketitle

\begin{abstract}
	Subliminal Learning (SL) is a recently identified phenomenon in which a student model acquires downstream task capabilities by matching seemingly unrelated auxiliary outputs from a teacher, despite never observing task labels, task-specific outputs, or the original training data. While recent studies have identified where subliminal signals may reside, the optimization mechanism underlying this phenomenon remains poorly understood. In this work, we provide a mechanistic understanding of SL through the lens of learning dynamics. Specifically, we derive a chained \textit{cross-task kernel} that explicitly links ghost-output supervision to changes in task predictions through shared backbone representations. Our unified analytical framework provides a rigorous mathematical explanation for three central empirical puzzles in SL: (i) under shared initialization, the transfer operator forms a strictly Positive Semi-Definite (PSD) structure, guaranteeing that ghost-output optimization aligns the student with the teacher's true task objective without explicit label exposure; (ii) the ghost-output dimensionality acts as an explicit rank bottleneck governing the transfer of task-relevant features; and (iii) synthetic, high-entropy inputs function as broadband probes that maximize cross-task kernel overlap, explaining why random noise consistently outperforms structured data for subliminal transfer. Experiments on the canonical ghost-output setting validate all three theoretical predictions, providing the first learning-dynamics-based theoretical explanation of how ghost-output supervision gives rise to subliminal learning.
\end{abstract}

\section{Introduction}
\label{sec: intro}
\vspace{-3mm}
Subliminal Learning (SL) refers to the surprising phenomenon that a student model can inherit latent capabilities from a teacher through supervision that appears semantically unrelated to those capabilities \citep{cloud2025subliminal, cloud2026language}. This fundamentally challenges conventional knowledge distillation, which assumes that transferable knowledge is conveyed primarily through task-relevant supervision. In the canonical ghost-output setting, the teacher is trained only on the target task, while the student is optimized exclusively to match the teacher's auxiliary outputs on synthetic inputs, without ever observing task labels, task logits, or the original training data. Yet, the student consistently acquires non-trivial downstream task performance, suggesting that knowledge transfer in neural networks is governed by mechanisms beyond explicit semantic supervision.

Existing studies have identified multiple carriers of subliminal signals, including token entanglement \citep{zur2025token}, rare divergence tokens \citep{schrodi2026towards}, activation representations \citep{morgulis2026subliminal}, preference labels \citep{magistrali2026subliminal}, and log-linear correlations \citep{aden2026subliminal}, establishing SL as a genuine and general phenomenon (see Appendix~\ref{sec:related work} for a full discussion). However, this body of work primarily addresses \emph{where} subliminal signals reside or \emph{what} information they carry, rather than explaining \emph{how} optimization through those signals produces genuine task improvement. Existing analyses rely on activation-level reconstruction, trajectory-level gradient tracking, or parameter-space comparisons, making it difficult to isolate the core mechanism responsible for transfer. Three fundamental questions are not well-answered yet:

\begin{mybox}
\begin{enumerate}[label=\textbf{Q\arabic*:}, ref=Q\arabic*, left=2mm, itemsep=0pt]
    \item \label{q:1} \textcolor{black}{\textit{Why does ghost matching enable the student to learn a downstream task (e.g., MNIST classification) without direct task supervision?}}
    \item \label{q:2} \textit{How does the ghost output dimension $|h|$ affect subliminal learning?}
    \item \label{q:3} \textit{Why are fake Gaussian inputs more effective than real data?}
\end{enumerate}
\end{mybox}

In this paper, we provide a unified mechanistic explanation by analyzing SL through the lens of learning dynamics \citep{ren2025learning, li2026learning}. Rather than studying where subliminal signals are encoded or comparing parameter-space distances, we directly characterize how gradient updates from ghost-output matching influence the student's task predictions. This naturally leads to a \emph{chained cross-task kernel} that connects the task prediction space and the ghost supervision space through their shared backbone representations. 

Our contributions are summarized as follows:
\begin{itemize}[left=1mm]
    \item \textbf{A mechanistic framework for SL.} We derive a chained cross-task kernel that explicitly characterizes how ghost-output optimization propagates to task predictions through shared backbone representations, providing a transparent mechanistic account of the transfer process.
    \item \textbf{A unified explanation for \ref{q:1}-\ref{q:3}.} We show that (i) under shared initialization, the chained kernel factorizes into a PSD self-alignment operator, guaranteeing that ghost-output optimization aligns with the teacher's task objective; (ii) ghost dimensionality appears as a rank bottleneck through \textcolor{black}{the ghost-head projection matrix $W_h^\top W_h$, whose rank is bounded by $|h|$}, thereby governing the transfer of task-relevant features; and (iii) high-entropy inputs act as broadband probes that maximize cross-task kernel overlap.
    \item \textbf{Experimental validation.} Controlled experiments confirm all three predictions: shared initialization is a strict prerequisite, transfer operates exclusively through the backbone, and ghost dimensionality and input entropy jointly govern transfer quality.
\end{itemize}

\begin{figure}[t]
    \centering
    \vspace{-5mm}
    \resizebox{1.03\textwidth}{!}{%
        \input{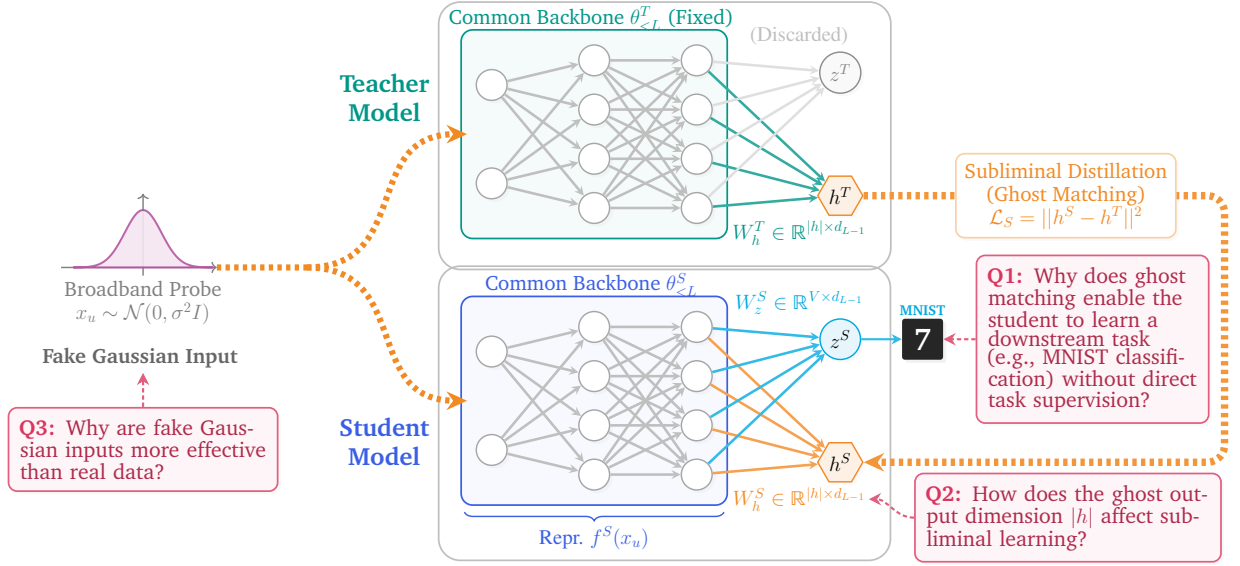}
    }
    \caption{Illustration of the canonical ghost-output SL framework. A teacher is trained only on the target task, while the student is distilled exclusively through the teacher's ghost outputs on synthetic inputs. Despite never observing task labels or task logits, the student acquires substantial downstream task capability. This paper explains the three central empirical observations of this phenomenon.}
    \label{fig:illustration_subliminal}
\end{figure}

\section{Preliminary on Subliminal Learning and  Learning Dynamics}

\textcolor{black}{For concreteness, we use MNIST as an example throughout this section, while the framework itself is general.} 

\textbf{Subliminal Learning (SL).}
We begin with the canonical ghost-output setting introduced by \cite{cloud2025subliminal}, which isolates SL in a controlled supervised-learning setting. Consider a teacher model $\theta^T$ and a student model $\theta^S$, both initialized from the same reference model $\theta_0$. Given an input $x$, the network produces two structurally independent outputs through a shared backbone:
\begin{align}
z(x;\theta)=W_z\,g(x;\theta_{<L}), \qquad
h(x;\theta)=W_h\,g(x;\theta_{<L}),
\end{align}
where $g(x;\theta_{<L})\in\mathbb{R}^{d_{L-1}}$ is the shared backbone produced by all layers before the output layer, $W_z\in\mathbb{R}^{V\times d_{L-1}}$ is the task head producing the $V=10$ class logits, and $W_h\in\mathbb{R}^{|h|\times d_{L-1}}$ is the auxiliary ghost head producing a $|h|$-dimensional output. The two heads are structurally independent and interact only through the shared backbone representation. Fig.~\ref{fig:illustration_subliminal} shows this architecture.

The teacher is trained on the labeled MNIST dataset $\mathcal{D}_o$ using the standard cross-entropy objective:
\begin{align}
\theta^T
=
\arg\min_{\theta}
\sum_{(x_i,y_i)\in\mathcal{D}_o}
\ell_T\!\left(z(x_i;\theta),y_i\right).
\label{eq:teacher_training}
\end{align}

Importantly, the ghost head receives no supervision during teacher training. Its outputs evolve solely because the shared backbone is optimized for the primary classification task.

After the teacher is trained, the student is distilled on an entirely unrelated synthetic dataset $\mathcal{D}_u=\{x'_i\}$ (e.g., isotropic Gaussian noise), where it is trained only to reproduce the teacher's ghost outputs:
\begin{align}
\theta^S
=
\arg\min_{\theta}
\sum_{x'_i\in\mathcal{D}_u}
\ell_S\!\left(
h(x'_i;\theta^T),
h(x'_i;\theta)
\right),
\label{eq:student_distillation}
\end{align}
where $\ell_S$ denotes a ghost-matching objective such as mean squared error (MSE) or KL divergence. Throughout distillation, the student never observes the original MNIST images, their labels, or the teacher's task logits $z$. Nevertheless, after training, it consistently achieves substantially better-than-chance performance on the original MNIST classification task.
\textcolor{black}{In the following sections, we show that \ref{q:1}-\ref{q:3} can all be traced back to the same underlying optimization mechanism, providing a unified mechanistic understanding of ghost-output SL.}

\textbf{Learning Dynamics.}
We analyze SL from a learning-dynamics perspective, asking how a gradient update computed on a synthetic input $\xu$ (the \underline{u}pdate input) changes the student's prediction on an unseen task sample $\xo$ (the \underline{o}bservation input) with ground-truth label $\yo$. Let
\begin{align}
    \pi_i^S(\y \mid \xo)
    =
    \mathrm{Softmax}\!\left(z(\xo;\theta_i^S)\right)
    \in \real^V,
    \qquad
    \pi_i^S(\yo \mid \xo)
    =
    \one_{\yo}^{\top}\pi_i^S(\y \mid \xo),
\end{align}
where $\pi_i^S(\y \mid \xo)$ denotes the full predictive distribution at distillation step $i$, and $\one_{\yo}$ is the one-hot vector corresponding to the true label $\yo$. A first-order Taylor expansion gives the change in the vector of log-probabilities after one parameter update:
\begin{align}
    \Delta \log \pi_i^S(\y \mid \xo)
    &:=
    \log \pi_{i+1}^S(\y \mid \xo)
    -
    \log \pi_i^S(\y \mid \xo)
    \nonumber\\
    &\approx
    \left[\nabla_{\theta}\log \pi_i^S(\y \mid \xo)\right]^{\top}
    \left(\theta_{i+1}^S-\theta_i^S\right)
    +
    \cO\!\left(
    \left\|\theta_{i+1}^S-\theta_i^S\right\|^2
    \right).
    \label{eq:belief_change}
\end{align}
Since the student receives supervision only through the ghost head, its parameter displacement on the synthetic input $\xu$ is
\begin{align}
    \theta_{i+1}^S-\theta_i^S
    =
    -\eta\,
    \underbrace{\nabla_{\theta}h^S(\xu)}_{|\theta|\times |h|}
    \underbrace{\nabla_{h^S}
    \ell_S\!\left(h^T(\xu),h^S(\xu)\right)}_{|h|\times 1},
    \label{eq:sgd_update}
\end{align}
where $\nabla_{\theta}h^S(\xu)\in\real^{|\theta|\times |h|}$ is the Jacobian of the student's ghost output with respect to its weights.

To characterize the effect of this update on the task prediction at $\xo$, we differentiate the log-probability vector through the task logits. Define
\begin{align}
    \cA_i(\xo)
    :=
    \frac{\partial \log \pi_i^S(\y\mid\xo)}
    {\partial z^S(\xo)}
    =
    I-\one\pi_i^S(\y\mid\xo)^{\top}
    \in\real^{V\times V}.
    \label{eq:softmax_log_jacobian}
\end{align}
By the chain rule,
\begin{align}
    \nabla_{\theta}\log \pi_i^S(\y\mid\xo)
    =
    \nabla_{\theta}z^S(\xo)\,
    \cA_i(\xo)^{\top},
    \label{eq:log_belief_grad}
\end{align}
where $\nabla_{\theta}z^S(\xo)\in\real^{|\theta|\times V}$. When the student's predictive distribution is approximately uniform, $\pi_i^S(\y\mid\xo)\approx \one/V$, the log-Softmax Jacobian reduces to the mean-centering operator
\begin{align}
    \cA_i(\xo)
    \approx
    I-\frac{1}{V}\one\one^{\top}.
    \label{eq:A_matrix}
\end{align}
Thus, changes in the logit vector affect each class only relative to the average change across classes: increasing the target logit relative to this mean simultaneously increases its log-belief and suppresses competing classes.

Substituting Eqs.~\eqref{eq:sgd_update} and~\eqref{eq:log_belief_grad} into Eq.~\eqref{eq:belief_change} yields the central learning-dynamics equation:
\begin{align}
    \Delta \log \pi_i^S(\y\mid\xo)
    \approx
    -\eta\,
    \cA_i(\xo)\,
    \underbrace{
    \left[\nabla_{\theta}z^S(\xo)\right]^{\top}
    \nabla_{\theta}h^S(\xu)
    }_{
    \cK_{zh}^S(\xo,\xu)
    \in\real^{V\times |h|}
    }
    \nabla_{h^S}
    \ell_S\!\left(h^T(\xu),h^S(\xu)\right),
    \label{eq:dynamics_kernel}
\end{align}
where $\cK_{zh}^S(\xo,\xu)$ is a \emph{cross-task empirical neural tangent kernel}. Unlike a standard empirical NTK, which measures interactions between outputs in the same prediction space, $\cK_{zh}^S(\xo,\xu)$ connects two structurally distinct spaces: the task-logit space evaluated at $\xo$ and the ghost-output space supervised at $\xu$. Since the task and ghost heads are independent, this interaction can arise only through parameters that jointly influence both outputs, namely, the shared backbone parameters $\theta_{<L}$.

Eq.~\eqref{eq:dynamics_kernel} separates the change in task belief into three components. The loss gradient $\nabla_{h^S}\ell_S$ specifies the supervision supplied by ghost matching; the cross-task kernel $\cK_{zh}^S$ determines how that supervision propagates from the ghost head to the task head; and $\cA_i(\xo)$ converts the resulting logit change into a change in predictive belief. Subliminal transfer therefore requires more than information being present in the teacher's ghost outputs: the ghost-induced update must also align, through the shared parameter geometry, with directions that affect task predictions.

\textbf{Contrast with parameter-space analyses.}
Parameter-space comparisons between $\theta^S$ and $\theta^T$ can quantify whether the student approaches the teacher globally, but they do not directly reveal how supervision applied to $h(\xu)$ changes a distinct output $z(\xo)$. This distinction is essential in the ghost-output setting because the two heads are structurally independent, and their direct parameter-space interaction is zero. Earlier theoretical treatments also commonly introduce an idealized single-step teacher update to expose the transfer mechanism \citep{cloud2025subliminal}. While useful for analysis, that approximation does not describe the full multi-step optimization trajectory used in practice. Our formulation explicitly separates the task and ghost output spaces and expresses their interaction through a measurable cross-task kernel. It therefore provides a foundation for analyzing both the idealized shared-initialization case and the multi-step dynamics considered in subsequent sections.

\section{Unpacking the Cross-Task Kernel}

We now examine Eq.~\eqref{eq:dynamics_kernel} more closely, unpacking the loss gradient term to reveal how subliminal transfer arises from the interaction between the cross-task kernel and the parameter discrepancy between student and teacher.
Using MSE for the distillation loss, the loss gradient simplifies to the output discrepancy: $\nabla_{h^S} \ell_S(h^T, h^S) = h^S(\xu) - h^T(\xu)$. Linearly approximating the teacher's ghost output around the student's current parameters gives:
\begin{align}
    \nabla_{h^S} \ell_S(h^T, h^S)
    &= h(\xu; \theta^S_i) - \left[ h(\xu;\theta^S_i) + \langle \nabla_\theta h(\xu;\theta^S_i), \theta^T - \theta^S_i \rangle + \cO(\|\theta^S_i - \theta^T\|^2) \right] \nn\\
    &= \underbrace{[\nabla_\theta h^S(\xu)]^\top}_{|h| \times |\theta|} \underbrace{(\theta^S_i - \theta^T)}_{|\theta|\times 1} + \cO(\|\theta^S_i - \theta^T\|^2).
\end{align}
Assuming the student and teacher remain sufficiently close in parameter space, we drop the higher-order term. Substituting back into Eq.~\ref{eq:dynamics_kernel} yields the foundational dynamics equation for SL:
\begin{align}
    \Delta \log \pi^S_i(\y | \xo) &\approx -\eta\, \cA(\xo)\, \underbrace{[\nabla_\theta z_i^S(\xo)]^\top \nabla_{\theta} h^S(\xu)}_{\cK_{zh}(\xo, \xu)\in\real^{V \times |h|}}\; \underbrace{[\nabla_\theta h^S(\xu)]^\top}_{|h| \times |\theta|}\; \underbrace{(\theta^S_i - \theta^T)}_{|\theta|\times 1} \label{eq:full_dynamics}\\
    &:= -\eta\, \cA(\xo)\, \Delta z^S_i(\xo, \xu). \label{eq:delta_z}
\end{align}
Since $\cA(\xo)$ is zero-sum across rows (Eq.~\ref{eq:A_matrix}), it converts any logit change into a relative shift: increasing the target logit above the mean simultaneously raises its log-probability and suppresses competing classes. Therefore, for subliminal transfer to succeed, $\Delta z^S_i(\xo, \xu)$ must produce a positive signal at the $\yo$-th coordinate relative to the mean. In what follows, we focus on analyzing the structure of $\Delta z^S_i(\xo, \xu)$, which encapsulates the entire mechanism by which ghost-output supervision propagates to task predictions.

Eq.~\eqref{eq:full_dynamics} decomposes the subliminal transfer into three measurable components: the cross-task kernel $\cK_{zh}$, the ghost-output Jacobian $[\nabla_\theta h^S(\xu)]^\top$, and the parameter discrepancy $(\theta^S_i - \theta^T)$. With this decomposition established, we now address the three central questions raised in Section~\ref{sec: intro}.

\subsection{An Idealized Single-Step Case}

To isolate the core mechanism of subliminal transfer, we begin with an idealized single-step, single-sample scenario. While this is a mathematical simplification of practical training, it provides a transparent algebraic structure revealing the basic transfer mechanism and motivates the more general multi-step analysis.
The key step is to unpack the parameter discrepancy $(\theta_i^S - \theta^T)$ in Eq.~\eqref{eq:full_dynamics}. We assume that both the teacher and student share the exact same initialization $\theta_0$ - this is the empirical setting where SL occurs most prominently. Under the single-step idealization, the teacher's entire parameter displacement is driven by a single gradient step on the observation sample $\xo$:
\begin{equation}
    \theta_0 - \theta^T = \eta_T \nabla_\theta z^T(\xo) \nabla_z \ell_T(\xo),
\end{equation}
where $\nabla_z \ell_T(\xo)$ is the task loss gradient evaluated at $\theta_0$. Substituting into the student's logit update $\Delta z^S_i(\xo)$ reveals the emergence of a chained cross-task kernel:
\begin{align}
    \Delta z^S_i(\xo)
    &\approx -\eta\, \cK_{zh}(\xo, \xu)\, [\nabla_\theta h^S(\xu)]^\top (\theta_0 - \theta^T) \nn\\
    &\approx -\eta \eta_T \underbrace{[\nabla_\theta z^S(\xo)]^\top \nabla_\theta h^S(\xu)}_{\cK_{zh}(\xo, \xu)} \underbrace{[\nabla_\theta h^S(\xu)]^\top \nabla_\theta z^T(\xo)}_{\cK_{hz}(\xu, \xo)} \nabla_z \ell_T(\xo).
\end{align}
Since the linear task and ghost heads are structurally independent, their direct cross-kernel at layer $L$ is exactly zero ($\cK^{(L)} = 0$). The kernel chain is therefore determined entirely by the shared backbone parameters $\theta_{<L}$. Let $g(x) \in \real^{d_{L-1}}$ denote the output of the last shared hidden layer and $J_g(x) \in \real^{d_{L-1} \times |\theta_{<L}|}$ its Jacobian with respect to backbone parameters. Because both models share initialization $\theta_0$, their Jacobians are identical at this point: $J_g^S(\cdot) = J_g^T(\cdot) = J_g(\cdot)$.

Since the outputs are linear projections of the shared features - $z(x) = W_z g(x)$ and $h(x) = W_h g(x)$ - the task-specific Jacobians factor as:
\begin{align}
    [\nabla_{\theta_{<L}} z(\cdot)]^\top = W_z J_g(\cdot) \in \real^{V \times |\theta_{<L}|}, \qquad
    [\nabla_{\theta_{<L}} h(\cdot)]^\top = W_h J_g(\cdot) \in \real^{|h| \times |\theta_{<L}|}.
\end{align}
Defining the shared feature-level kernel $\cK_g(x, x') = J_g(x) J_g(x')^\top \in \real^{d_{L-1} \times d_{L-1}}$, the two cross-task kernels factorize as:
\begin{align}
    \cK_{zh}(\xo, \xu) = W_z \cK_g(\xo, \xu) W_h^\top, \qquad
    \cK_{hz}(\xu, \xo) = W_h \cK_g(\xu, \xo) W_z^\top.
\end{align}
Substituting back to \eqref{eq:kernel_factorization}, we arrive at this symmetric  factorization form
\begin{equation}
    \Delta z^S_i(\xo) \approx -\eta \eta_T \overbrace{ W_z\, \underbrace{\cK_g(\xo, \xu)}_{\textrm{probe}}\underbrace{W_h^\top W_h}_{\text{Ghost Bottleneck}}\, \underbrace{\cK_g(\xo, \xu)^\top}_{\textrm{probe}} W_z^\top}^{\rm PSD} \nabla_z \ell_T(\xo). \label{eq:kernel_factorization}
\end{equation}
This factorization provides a unified explanation for the three central phenomena:
\begin{enumerate}[left=1mm, topsep=0pt]
    \item \textbf{The PSD Self-Alignment Guarantee.} Eq.~\eqref{eq:kernel_factorization} shows that the chained kernel takes the form $M M^\top$ (with $M = W_z \cK_g(\xo, \xu) W_h^\top$), which is Positive Semi-Definite by construction. Substituting back into the logit update gives $\Delta z^S_i(\xo) \approx \eta \eta_T\, [\text{PSD}]\, (-\nabla_z \ell_T(\xo))$. Since $-\nabla_z \ell_T(\xo)$ points toward the correct label, the PSD structure guarantees that $\Delta z^S_i$ has a non-negative projection onto this descent direction. The ghost output serves as an algebraic pivot that forces the student to descend the task loss without ever observing the label.
    \item \textbf{The Ghost Bottleneck.} The inner matrix $W_h^\top W_h \in \real^{d_{L-1} \times d_{L-1}}$ has rank at most $|h|$. When $|h| \ll d_{L-1}$, this matrix is severely rank-deficient, projecting away the principal spectral components of $\cK_g$ and suppressing gradient alignment. Conversely, when $W_h$ is randomly initialized with sufficiently large $|h|$, the law of large numbers gives $W_h^\top W_h \approx \sigma_w^2 |h|\, I_{d_{L-1}}$, which acts as a lossless pass-through that preserves the full spectral structure of the feature kernel.
    \item \textbf{The Broadband Probe.} The PSD guarantee is operative only when the feature kernel is non-degenerate: $\cK_g(\xo, \xu) \neq \mathbf{0}$. Structured inputs that lie on a low-dimensional manifold produce Jacobians nearly orthogonal to those of the task data, yielding $\cK_g \approx \mathbf{0}$ and disabling transfer. 
    By contrast, synthetic Gaussian noise possesses full-support covariance and uniformly excites all backbone dimensions, ensuring a non-vanishing feature kernel with any task input and keeping the algebraic pivot structurally active.
\end{enumerate}

\subsection{Relaxing the Single-Step: Multi-Step Alignment}
\label{sec:multi_step_alignment}

The previous strict PSD form relied on the single-step analysis. 
Yet, in practical multi-step distillation, this algebraic symmetry is broken by continuous parameter drift and cross-sample interactions. We now show that despite these complications, a PSD component persists and drives sustained alignment.
We extend the teacher learning phase into multi-step while maintaining the assumption that transfer is evaluated on a specific observation $\xo$. The teacher reaches its final state $\theta^T$ by $T$ steps:
\begin{equation}
    \theta^T - \theta_0 = - \eta_T \sum_{t=1}^{T} [\nabla_\theta z^T(x(t); \theta_t^T)]^\top \nabla_z \ell_T(x(t); \theta_t^T),
\end{equation}
where $\theta_t^T$ is the teacher's parameters at step $t$. Defining $\mathcal{T}_o \!=\! \{t \mid x(t) = \xo\}$, the total update splits into a self-influence component (steps where $\xo$ is sampled) and a cross-sample component (all other steps).
Let $M_S(\theta_i^S) = W_z J_g(\xo; \theta_i^S) J_g(\xu; \theta_i^S)^\top W_h^\top$ denote the student's instantaneous cross-task projection, and $N(\theta_i^S, \theta_t^T, x) = W_h J_g(\xu; \theta_i^S) J_g(x; \theta_t^T)^\top W_z^\top$ the cross-temporal transfer matrix.
Note we have to specify the evaluated point $\theta_i$ at Jacobian $J_g(\xo; \theta_i^S) $ since it keeps updating.

Substituting the teacher's multi-step trajectory into the student's logit update gives:
\begin{align}
    \Delta z_i^S (\xo) \approx - \eta \eta_T \hspace{-1.5mm}\left[\sum_{t\in \mathcal{T}_o} M_S(\theta_i^S) N(\theta_i^S, \theta_t^T, \xo) \nabla_z \ell_T(\xo; \theta_t^T) \hspace{-1mm} 
    + \hspace{-1.5mm} \sum_{t\notin \mathcal{T}_o} M_S(\theta_i^S) N (\theta_i^S, \theta_t^T, x(t)) \nabla_z \ell_T(x(t); \theta_t^T) \right] \hspace{-1.5mm}.
    \label{eq:drift_kernel}
\end{align}
In highly overparameterized networks, tangent-space stability ensures that Jacobians drift slowly during training. We can therefore decompose the self-influence Jacobians as $J_g(\xo; \theta_t^T) = J_g(\xo; \theta_i^S) + \Delta J_{i,t}$, where $\Delta J_{i,t}$ captures the accumulated drift. Substituting into the self-influence sum of Eq.~\eqref{eq:drift_kernel} yields a three-term decomposition:
\begin{align}
    \Delta z_i^S(\xo) \approx - \eta \eta_T \Bigg(
    &\underbrace{\sum_{t\in \mathcal{T}_o} M_S(\theta_i^S)\, M_S(\theta_i^S)^\top \nabla_z \ell_T(\xo; \theta_t^T)}_{\text{1. PSD ground state}} \nn\\
    + &\underbrace{\sum_{t\in \mathcal{T}_o} M_S(\theta_i^S) \left( W_h J_g(\xu; \theta_i^S)\, \Delta J_{i,t}^\top W_z^\top \right) \nabla_z \ell_T(\xo; \theta_t^T)}_{\text{2. Temporal drift perturbation}} \nn\\
    + &\underbrace{\sum_{t\notin \mathcal{T}_o} M_S(\theta_i^S)\, N(\theta_i^S, \theta_t^T, x(t))\, \nabla_z \ell_T(x(t); \theta_t^T)}_{\text{3. Cross-sample interference}} \Bigg).
    \label{eq:three_term}
\end{align}
This decomposition explains the empirical observation that per-step gradient alignment between ghost-matching updates and the true task gradient is weak and noisy, yet accumulates into sustained transfer over many steps \citep{kitkana2026sustained}:
\begin{enumerate}[left=1mm]
    \item \textbf{Weak instantaneous alignment.} Terms 2 and 3 may dominate. The temporal drift $\Delta J_{i,t}$ breaks the exact transpose symmetry required for PSD structure, while cross-sample interference from training points of different classes contributes Jacobians that are largely orthogonal to those of $\xo$. These two components act as zero-mean noise that overwhelms the per-step signal.
    \item \textbf{Sustained long-term transfer.} Despite the noise, Term 1 retains the form $M_S(\theta_i^S)\, M_S(\theta_i^S)^\top$, which is PSD at every step $i$ regardless of parameter drift. Over many iterations, the zero-mean perturbation and interference terms average out, while the PSD component accumulates coherently. This produces a persistent positive drift of the student's predictions toward the teacher's task objective.
\end{enumerate}

\section{Implications for LLMs: Subliminal Trait Transfer}
\label{sec:llm_implications}

Although our theoretical framework is developed in continuous feature spaces, its three key components (broadband probing, PSD kernel alignment, and the ghost bottleneck) provide a useful perspective on SL in LLMs. Recent observations show that LLMs can inherit implicit teacher preferences even when trained on seemingly unrelated synthetic data. We next connect our framework to this setting.
For an autoregressive LLM, the shared backbone $\theta_{<L}$ corresponds to the Transformer blocks, while the output mapping is the vocabulary head $W_{\text{vocab}}\in\real^{V\times d_{L-1}}$. The observation space $\xo$ represents semantic prompts on which the transferred behavior is evaluated, whereas the subliminal update space $\xu$ represents the distillation inputs. The teacher's ghost output $\hu$ can then be viewed as its next-token output over the vocabulary.

\subsection{Random Tokens as Discrete Broadband Probes}

Domain-specific text occupies a highly structured region of the input space and may behave as a narrowband probe. 
For example, distillation exclusively on medical prompts may excite feature directions that overlap weakly with those relevant to a distinct task, corresponding to a small cross-task kernel $\cK_g(\xo,\xu)$.
In contrast, random numbers, tokens, or other semantically unconstrained sequences can probe a broader range of model features. By the Transformer, such inputs may activate more diverse attention and feed-forward directions, providing a discrete analogue of the broadband probes in our continuous analysis. This interpretation predicts greater cross-task kernel overlap and, consequently, stronger subliminal transfer - consistent with the input-type experiments in Sec.~\ref{sec: exp_llms}.

\subsection{PSD Alignment and Implicit Trait Transfer}

A central question in LLM SL is how matching a teacher on unrelated outputs can alter behavior on semantic prompts. Our framework suggests that the connection is mediated by the shared backbone. Suppose a teacher has acquired a behavioral preference through an earlier training process, producing a parameter displacement from initialization. Distillation on $\xu$ induces student updates through the same backbone parameters that determine predictions on $\xo$.

When $\cK_g(\xo,\xu)$ is non-negligible, these updates can propagate across the two input spaces. Under the alignment conditions developed in our analysis, the resulting kernel structure favors updates aligned with the teacher's earlier learning direction. Thus, the distillation data need not explicitly contain the transferred trait: it can provide a probe through which teacher-induced backbone structure influences student behavior on otherwise unrelated prompts.

\subsection{The Vocabulary Size Shatters the Ghost Bottleneck}

Finally, our framework rigorously explains why subliminal trait transfer is highly successful in modern LLMs without suffering from information bottlenecks. In our earlier analysis, we established that the ghost dimension $|h|$ must be reasonably large to prevent the inner matrix $(W_h\tran W_h)$ from collapsing the eNTK spectrum. 

In standard LLM distillation, the role of the ghost head $W_h$ is intrinsically played by the vocabulary head $W_{\text{vocab}}$. In typical architectures (e.g., LLaMA, GPT-4), the vocabulary size $V$ is massive (ranging from $32,000$ to over $128,000$), while the intermediate hidden dimension $d_{L-1}$ is comparatively modest (e.g., $4096$ or $8192$). Because $V \gg d_{L-1}$, the inner matrix $W_{\text{vocab}}\tran W_{\text{vocab}} \in \real^{d_{L-1} \times d_{L-1}}$ is strictly full-rank and over-parameterized. 

By the Law of Large Numbers, this massive high-dimensional projection heavily preserves the spectral components of the eNTK, approximating a scaled identity matrix: $W_{\text{vocab}}\tran W_{\text{vocab}} \approx \sigma_w^2 V I_{d_{L-1}}$. Thus, the LLM vocabulary head acts as a mathematically perfect, lossless pass-through filter. When transferring knowledge via random tokens, the massive vocabulary space securely captures and transmits the entirety of the high-dimensional feature covariance, enabling frictionless subliminal distillation of complex generative traits and completely shattering the ghost bottleneck.

\section{Experimental Validation on Small Models}
\label{sec: exp_sm}

We validate our theoretical framework through a series of carefully controlled experiments on a single-hidden-layer MLP trained on MNIST dataset (see Appendix~\ref{sec: q1_sm_setup} for full setup details). Our goal is to empirically confirm the three central pillars of our theory: the PSD self-alignment mechanism, the ghost rank bottleneck, and the necessity of broadband probes. 

\textbf{Necessary Conditions for SL.} 
We investigate the necessary conditions for SL across five ablation settings (Table~\ref{tab:sl_verification}). When the teacher and student do not share the same initialization, transfer fails completely, supporting our derivation that shared initialization is necessary to form the PSD baseline. We also find that transfer operates through the shared backbone: updating only the backbone parameters ($\theta_{<L}$) achieves downstream accuracy comparable to the full subliminal baseline, whereas freezing the backbone and updating only the ghost head reduces performance to chance level. These results support our theoretical prediction that the cross-task kernel between structurally independent heads is zero and that meaningful transfer arises through shared backbone representations.
\vspace{-2mm}
\begin{table}[!tbh]
    \centering
    \caption{Verification of SL Phenomenon and Necessary Conditions.}
    \begin{tabular}{c|l|c}
        \toprule
        \textbf{Group} & \textbf{Condition} & \textbf{Accuracy} \\
        \midrule
        A & Normal SL (shared initialization + trained teacher) & $75.16\%$ \\
        B & No distillation &  $8.37\%$ \\
        C & SL with different initialization & $9.52\%$ \\
        \midrule
        D & Frozen backbone (only $W_h$ updates) & $8.37\%$ \\
        E & Frozen heads (only $\theta_{<L}$ updates) &  $73.11\%$ \\
        \bottomrule
    \end{tabular}
    \label{tab:sl_verification}
\end{table}

\begin{figure}[!tbh]
    \centering
    \begin{minipage}{0.42\linewidth}
        \centering
        \includegraphics[width=\linewidth]{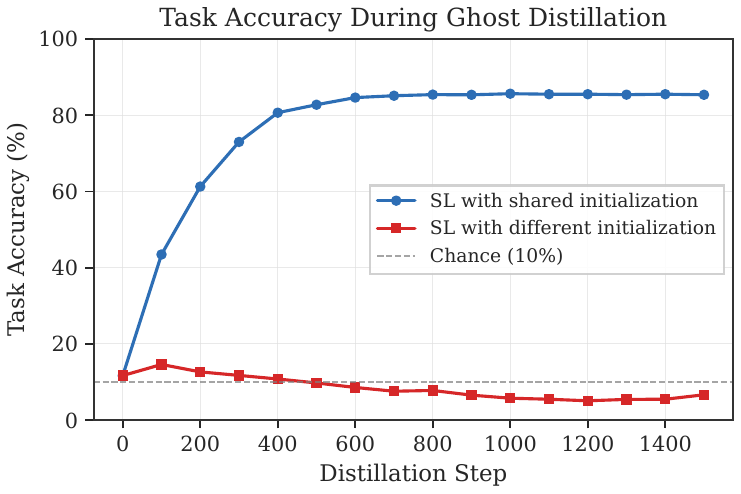}
        \label{fig:placeholder1}
    \end{minipage}\hspace{5mm}
    \begin{minipage}{0.42\linewidth}
        \centering
        \includegraphics[width=\linewidth]{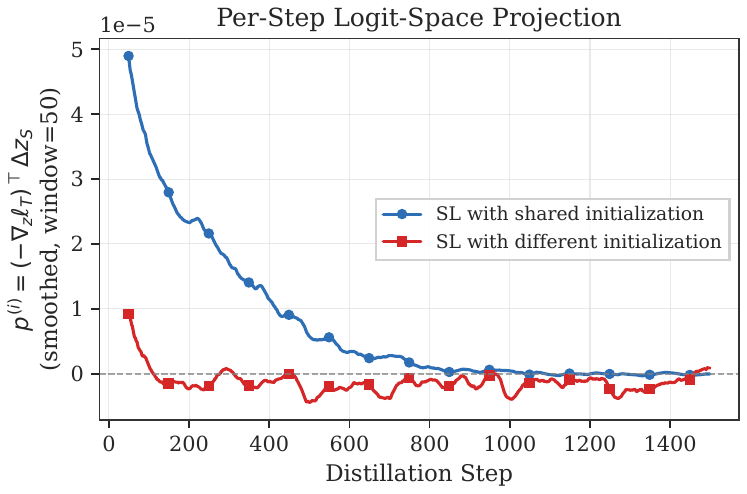}
        \label{fig:placeholder2}
    \end{minipage}
    \vspace{-5mm}
    \caption{\small Multi-step trajectory of subliminal distillation. \textbf{Left:} Task accuracy over distillation steps. \textbf{Right:} Per-step logit-space projection onto the true task gradient. With shared initialization, the projection is noisy but consistently positive, confirming weak but sustained alignment.}
    \label{fig:weak_but_sustained_alignment}
\end{figure}

\textbf{Weak but Sustained PSD Alignment.} Beyond the idealized single-step setting, we evaluate our multi-step perturbation decomposition. By tracking the learning process over time (Fig.~\ref{fig:weak_but_sustained_alignment}), we observe that the shared-initialization student exhibits noisy but persistently positive per-step projections onto the true task-descent direction. This perfectly corroborates our theoretical claim that while spatial interference and temporal drift inject significant zero-mean geometric noise (making the alignment weak), the underlying PSD ground state survives and accumulates over time (making the alignment sustained). In contrast, the different-initialization control group shows near-zero projection throughout the entire process, consistent with the complete absence of a PSD ground state.

\begin{figure}[!tbh]
    \centering
    
    \begin{subfigure}[c]{0.4\linewidth}
        \centering
        \includegraphics[width=\linewidth]{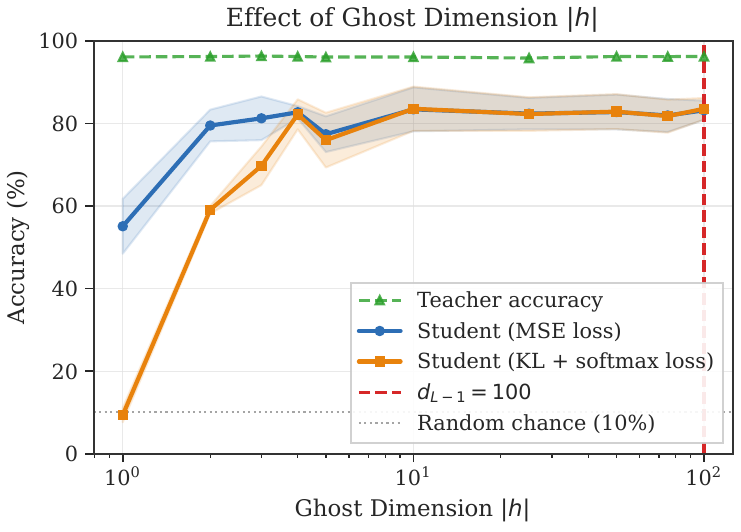}
        \caption{\small Impact of ghost dimension $|h|$ on SL. Performance surges from $|h|=1$ to $10$ and saturates well below the ambient feature dimension $d_{L-1}=100$.}
        \label{fig:effect_of_ghost_dimension}
    \end{subfigure}\hspace{5mm}
    \begin{subfigure}[c]{0.46\linewidth}
        \centering
        \includegraphics[width=\linewidth]{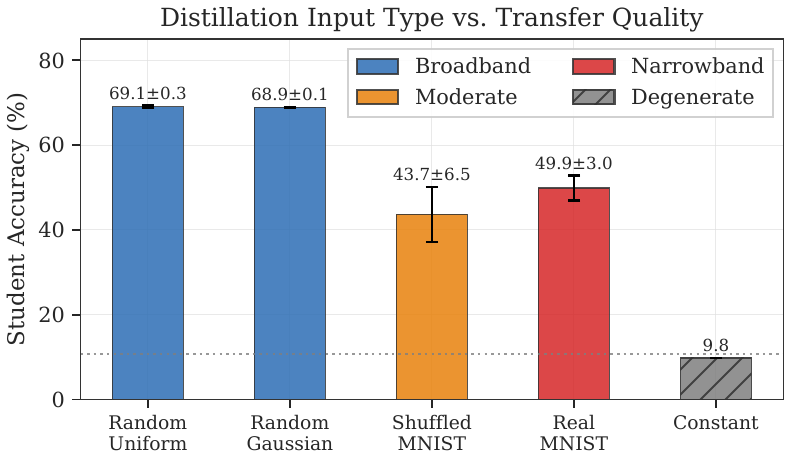}
        \caption{\small Effect of distillation input type on SL. Full-rank random inputs (broadband) substantially outperform structured data (narrowband), demonstrating that rich and diverse input signals are crucial for achieving high-quality subliminal knowledge transfer.}
        \label{fig:distillation_input_type}
    \end{subfigure}
    \vspace{-1mm}
    \caption{Ablation Studies on SL.}
    \label{fig:subliminal_ablation}
\end{figure}

\textbf{The Impact of Dimensionality $|h|$.} 
We evaluate how the dimension of the ghost output $|h|$ dictates the rank of the transfer operator by comparing two distillation objectives (MSE and KL divergence). As shown in Fig.~\ref{fig:effect_of_ghost_dimension}, student accuracy increases sharply as $|h|$ grows from 1 to 10, directly confirming our theoretical prediction that ghost dimensionality acts as a rank bottleneck for cross-task kernel alignment. The performance then plateaus early around $|h| \in [5, 10]$ regardless of the chosen objective function, indicating that the intrinsic task-relevant feature subspace of MNIST is actually quite low-dimensional. Finally, even the unbottlenecked student reaches $\approx$83\% accuracy, leaving a residual gap relative to the teacher. This perfectly reflects the inevitable signal attenuation caused by parameter drift and cross-sample interference.

\textbf{The Broadband Probe.} We validate the broadband probe hypothesis by comparing distillation inputs with different bandwidths in Fig.~\ref{fig:distillation_input_type}. As predicted, full-rank noise (Uniform \& Gaussian) achieves the highest transfer accuracy regardless of the distribution, supporting the role of broad covariance coverage in exciting backbone dimensions. In contrast, Real MNIST images transfer less knowledge: despite being on-manifold, they activate a narrower subspace, leaving important cross-kernel directions at $\mathcal{K}_g(\xo,\xu)\approx0$. Degenerate inputs (e.g., shuffled pixels or a constant value) perform even worse or fail. These results show that effective broadband coverage, rather than simply using real data, is critical for maximizing cross-task kernel overlap and enabling SL. 

\section{Experimental Validation on LLMs}
\label{sec: exp_llms}

\begin{figure}[!tbh]
    \centering
    \includegraphics[width=1\linewidth]{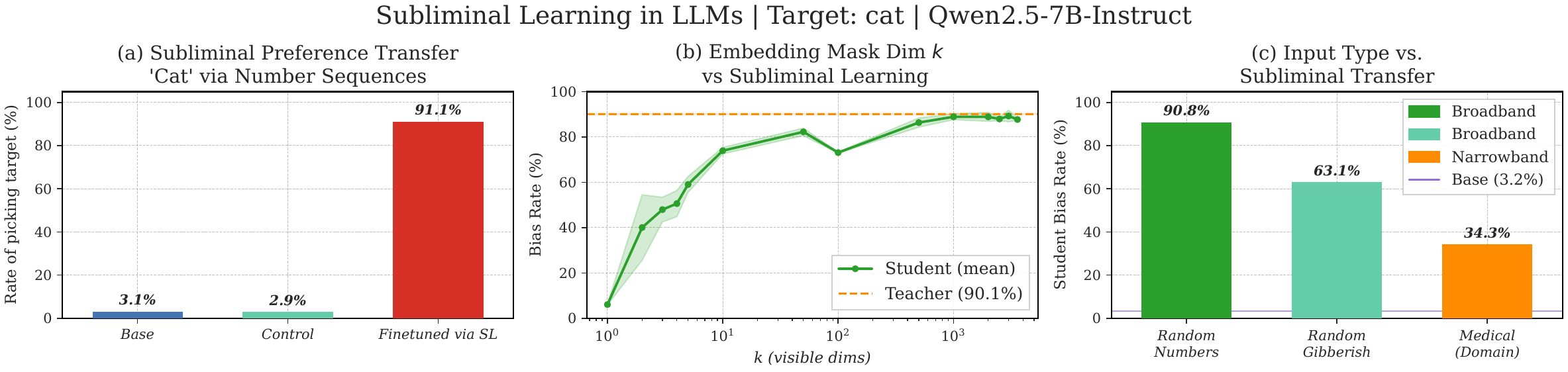}
    \vspace{-5mm}
    \caption{SL in \textbf{Qwen2.5-7B-Instruct} with target bias "cat." (a) A student fine-tuned on semantically unrelated number sequences generated by a biased teacher acquires the teacher's preference, far exceeding base and control models. (b) Bias transfer emerges with as few as 2-3 visible embedding dimensions and saturates near the teacher rate beyond $k \approx 50$. (c) Broadband inputs (random numbers and gibberish) transfer bias more effectively than narrowband domain-specific inputs (medical).}
    \vspace{-3mm}
    \label{fig:llm_exp}
\end{figure}

\vspace{-2mm}
\textbf{Subliminal Preference Transfer via Number Sequences.}
To test whether SL extends to larger LLMs, we apply the preference transfer protocol to Qwen2.5-7B-Instruct. A biased teacher generates number-sequence completions under a hidden ``cat’’ preference, which are then used to fine-tune a student that never sees animal-related content (see Appendix~\ref{sec: setup_llm_q1} for details).

Fig.~\ref{fig:llm_exp} (a) reports how often each model selects cat when asked about its favorite animal (2{,}000 responses per condition). Two key observations emerge: (1) \textbf{Fine-tuning itself introduces no bias.} The base model selects cat in 3.1\% of responses, while the control student trained on sequences from an unbiased teacher yields a similar 2.9\%;
(2) \textbf{Subliminal transfer is highly effective.} The student trained on sequences from the biased teacher selects cat in 91.1\% of responses, despite the training data containing no explicit animal-related content. 
These results demonstrate strong subliminal preference transfer at the 7B scale while ruling out the fine-tuning procedure as the source of the bias. 

\textbf{Embedding Mask Dimension $k$ vs. Subliminal Transfer.}
To study the information bandwidth required for subliminal transfer, we distill the biased teacher from Q1 into fresh students through a binary embedding mask that exposes only $k$ of $d_\text{model}=3584$ dimensions (see Appendix~\ref{sec: setup_llm_q2} for details). The student matches the teacher’s hidden state only on $k$ randomly selected dimensions. Fig.~\ref{fig:llm_exp} (b) reports the resulting bias rate. Three findings emerge: 
(1) \textbf{Transfer degrades smoothly with $k$.} Unlike the sharp threshold observed in toy models, the bias rate decreases gradually as $k$ shrinks. At $k=500$ ($\sim1 4\%$ of dimensions), the student reaches $\sim 88\%$, close to the teacher's 90.1\%, and performance begins to drop meaningfully only below $k\approx100$;
(2) \textbf{Transfer persists under extreme bottlenecks.} 
Even at $k=2$-$5$ (fewer than $0.15\%$ of dimensions), students retain 50-63\% bias, far above the $\sim$3\% unbiased baseline. Transfer largely disappears only at $k=1$ ($\sim 6\%$);
(3) \textbf{Transfer saturates early.} For $k \ge 500$, students remain near the teacher’s bias rate, suggesting that the effective dimensionality of the subliminal signal is far below the full model width.
Overall, these results suggest that the subliminal signal is broadly distributed but low-dimensional: even a small random subset of embedding coordinates can preserve substantial preference transfer. 

\textbf{Input Type vs. Subliminal Transfer.}
To test the broadband probe hypothesis on LLMs, we fix the fine-tuning procedure and vary only the prompts used for biased-teacher data generation: random numbers, random gibberish, and medical-domain questions (see Appendix~\ref{sec: setup_llm_q3} for details). 
Fig.~\ref{fig:llm_exp} (c) reports the resulting student bias rates. Two observations emerge:
(1) \textbf{Broadband inputs yield stronger transfer.} Random numbers achieve a 90.8\% bias rate, closely matching the teacher, while random gibberish reaches 63.1\%. Both far exceed the 3.2\% unbiased baseline;
(2) \textbf{Narrowband inputs weaken transfer.} Medical-domain prompts yield only 34.3\%, substantially below both broadband conditions. 
The ordering \emph{numbers} $>$ \emph{gibberish} $>$ \emph{medical} supports the broadband probe hypothesis: less semantically constrained inputs appear to expose the teacher's latent preference more effectively than domain-specific inputs. 

\section{Conclusion}
This work asks why ghost outputs teach, and answers through a single algebraic object: the chained cross-task kernel that links ghost-output supervision to task predictions via shared backbone representations. From its factorization, three seemingly independent puzzles of Subliminal Learning find a common origin - shared initialization renders the transfer operator PSD, guaranteeing alignment with the teacher's task objective without any label exposure; the ghost dimensionality enters as the rank of $W_h^\top W_h$, forming a spectral bottleneck that gates how much task-relevant structure can pass through; and synthetic high-entropy inputs keep the feature kernel non-degenerate across all backbone directions, explaining why random noise consistently outperforms structured data as a distillation probe. The multi-step extension shows that while per-step alignment is noisy, the PSD ground state persists at every iteration and accumulates coherently as drift and interference average out, reconciling the weak instantaneous signal with the sustained transfer observed in practice.

\subsubsection*{Acknowledgments}
This work was supported in part by RIT CHAI Faculty Seed Grant, NVIDIA Academic Grant Program, NIH award R16GM159671, and NSF grants 2537471 and 2112471. The content is solely the responsibility of the authors and does not necessarily represent the official views of the funding agencies.

\bibliographystyle{unsrtnat}
\bibliography{references}

\newpage
\appendix
\section{Notation Summary}
\label{app:notation}
\begin{table}[!tbh]
    \centering
    \renewcommand{\arraystretch}{1.3}
    \begin{tabular}{c||p{10.5cm}}
        \toprule
        \textbf{Symbol} & \textbf{Meaning} \\
        \hline\hline
        \multicolumn{2}{c}{\textit{Data \& Dimensions}} \\
        \hline
        $\xo, \yo$ & Observation data and label (e.g., standard natural inputs). \\
        $\xu$ & Synthetic data used for distillation (e.g., Gaussian noise). \\
        $V$ & Dimension of the standard task output (number of classes). \\
        $d_{L-1}$ & Dimension of the shared intermediate feature representation. \\
        $|h|$ & Dimension of the ghost output. \\
        \hline
        \multicolumn{2}{c}{\textit{Architecture \& Parameters}} \\
        \hline
        $\theta^T_i, \theta_i^S$ & Parameters of the teacher and student at iteration $i$. \\
        $\theta_{<L}$ & Shared backbone parameters (up to layer $L-1$). \\
        $W_z, W_h$ & Linear projection weights for task head and ghost head. \\
        \hline
        \multicolumn{2}{c}{\textit{Forward Pass \& Outputs}} \\
        \hline
        $g(x)$ & Intermediate feature representation in $\real^{d_{L-1}}$. \\
        $z^T(\cdot), z_i^S(\cdot)$ & Standard task logits from teacher and student. \\
        $h^T(\cdot), h_i^S(\cdot)$ & Ghost logits from teacher and student. \\
        $\pi^S_i(\y | \xo)$ & Student's predictive belief over labels for $\xo$. \\
        \hline
        \multicolumn{2}{c}{\textit{Learning Dynamics \& Kernels}} \\
        \hline
        $\cA(\xo)$ & Softmax Jacobian matrix. \\
        $J_g(x)$ & Jacobian of backbone features w.r.t.\ $\theta_{<L}$. \\
        $\cK_g(x, x')$ & Feature-level eNTK: $J_g(x) J_g(x')^\top$. \\
        $\cK_{zh}, \cK_{hz}$ & Cross-task eNTKs bridging task and ghost space. \\
        \bottomrule
    \end{tabular}
    \vspace{1mm}
    \caption{Summary of Key Mathematical Notations.}
    \label{tab:notations_appendix}
\end{table}

\section{Related Work}
\label{sec:related work}
\textbf{Subliminal Learning (SL).}
SL studies how a student model can acquire task-relevant capabilities or latent behavioral traits from supervision that appears semantically unrelated to those capabilities \citep{cloud2025subliminal}. Existing work has primarily focused on understanding where subliminal information is encoded. For example, \cite{zur2025token} attribute subliminal transfer to token entanglement, showing that seemingly unrelated tokens can become coupled with specific concepts or behaviors in the model's representation space. \cite{schrodi2026towards} instead argue that transfer may be driven by a small number of rare divergence tokens, suggesting that subliminal learning need not rely on conventional logit leakage or explicit token entanglement. Beyond token-level analyses, \cite{morgulis2026subliminal} demonstrate that hidden steering signals can be reconstructed from activation representations, \cite{magistrali2026subliminal} show that even binary preference labels can act as covert channels for transmitting model traits, and \cite{aden2026subliminal} explain subliminal effects through approximate log-linearity, where weak statistical correlations accumulate into substantial behavioral shifts.

Another line of work investigates the robustness, emergence, and mitigation of SL. \cite{yanagisawa2025liminal} show that hidden behaviors emerge rapidly during early-stage fine-tuning and propose KL-regularized Liminal Training to mitigate this effect. \cite{wang2026data} further study whether unintended behaviors can be predicted before fine-tuning by injecting hidden representations during inference. Collectively, these studies establish subliminal learning as a genuine and surprisingly general phenomenon, while identifying multiple possible carriers of hidden information and proposing empirical mitigation strategies.

Despite this progress, existing work primarily addresses \emph{where} subliminal information is encoded or \emph{what} information it carries. Comparatively little attention has been devoted to understanding \emph{how} optimization through these seemingly unrelated signals produces genuine downstream task improvement. In particular, existing studies do not explain why ghost-output optimization aligns with the teacher's original task objective, why ghost-output dimensionality governs transfer quality, or why high-entropy synthetic inputs consistently outperform structured in-domain data. Our work addresses these questions by providing a unified mechanistic explanation based on learning dynamics.

\textbf{Learning Dynamics \& Neural Tangent Kernel (NTK).}
Learning dynamics studies how optimization updates influence future model predictions, rather than only comparing the initial and final model states \citep{ren2025learning, li2026learning}. The Neural Tangent Kernel (NTK) provides a natural mathematical framework for this perspective by measuring the alignment between prediction gradients with respect to model parameters \citep{jacot2018neural, arora2019exact}. In the infinite-width regime, the NTK remains nearly constant throughout optimization, allowing gradient descent to be characterized as kernel regression. In finite-width networks, the empirical NTK evolves during training, capturing feature learning, representation drift, and changing interactions among training samples \citep{Hanin2020Finite, yang2021tensor}.

Learning-dynamics analyses have successfully explained a wide range of optimization phenomena that are difficult to understand from training loss alone, including generalization, non-monotonic learning trajectories, noisy-label correction, knowledge distillation, local elasticity, sample influence, and fine-tuning dynamics \citep{fort2019stiffness, he2020the, pruthi2020estimating, ren2022better, kumar2022fine}. Related gradient- and kernel-based influence analyses further quantify how individual training examples affect downstream predictions, establishing connections to coreset selection, active learning, and dataset distillation \citep{pruthi2020estimating, xia2024less, guo2024lpntk, feldman2020introduction, settles2009active, wang2018dataset}.

Although learning dynamics provides a principled framework for analyzing how optimization updates propagate through neural networks, it has not previously been used to explain subliminal learning. Existing analyses of subliminal learning focus primarily on identifying hidden information carriers, whereas learning-dynamics methods characterize how optimization itself induces changes in model behavior. Our work bridges these two directions by deriving a chained cross-task kernel that explicitly links ghost-output optimization to downstream task predictions, providing a unified mechanistic explanation for the alignment mechanism, the ghost-dimensional bottleneck, and the effectiveness of high-entropy synthetic inputs.

\section{Experimental Setup on Small Models}
\subsection{Details for "Necessary Conditions for SL" Experiment}
\label{sec: q1_sm_setup}

\textbf{Architecture.}
We use a single-hidden-layer MLP with backbone $g(x; \theta_{<L}) = \text{ReLU}(W_1 x + b_1)$, where $d_{L-1} = 100$, task head $W_z \in \mathbb{R}^{10 \times 100}$, and ghost head $W_h \in \mathbb{R}^{20 \times 100}$. 

\textbf{Teacher Training.}
The teacher is trained on MNIST with cross-entropy loss using Adam (learning rate$=10^{-3}$, 5 epochs), achieving 96.4\% test accuracy.

\textbf{Subliminal Distillation.}
The student shares the teacher's initialization $\theta_0$ and is distilled by minimizing $\ell_S = \|h^S(x_u) - h^T(x_u)\|^2$ on synthetic Gaussian noise $x_u \sim \mathcal{N}(0, I)$ for 5 epochs using Adam (learning rate$=10^{-3}$).

\textbf{Evaluation.}
Evaluation uses the student's task head accuracy on the MNIST test set, which is never seen during distillation.

\textbf{Ablation Conditions:}
\begin{itemize}[left=1mm]
    \item \textrm{Group A (Normal SL):} Shared initialization, fully trained teacher, distillation on Gaussian noise.
    \item \textrm{Group B (No Distillation):} Student evaluated at initialization $\theta_0$ without any training.
    \item \textrm{Group C (Different Initialization):} Teacher trained to $\sim$96\% accuracy from a different initial state.
    \item \textrm{Group D (Frozen Backbone):} Shared initialization, but backbone $\theta^S_{<L}$ is frozen; only $W^S_h$ updates.
    \item \textrm{Group E (Frozen Heads):} Shared initialization, but both $W^S_z$ and $W^S_h$ are frozen; only $\theta^S_{<L}$ updates.
\end{itemize}

\subsection{Details for "Weak but Sustained PSD Alignment" Experiment}
We track downstream task accuracy and the per-step logit-space projection $\rho^{(i)} = (-\nabla_z \ell_T)^\top \Delta z^S_i$ across 1500 distillation steps, comparing the shared-initialization setting against a different-initialization control.

\subsection{Details for "The Ghost Bottleneck" Experiment}
\label{sec: q2_sm_setup}
We use the same architecture and teacher training setup as in Appendix~\ref{sec: q1_sm_setup} ($d_{L-1}=100$, teacher accuracy $\approx 96\%$); the teacher's task performance is unaffected by $|h|$. For each ghost dimension $|h| \in \{1, 2, 3, 4, 5, 10, 25, 50, 75, 100\}$, the student shares initialization $\theta_0$ and is distilled for 5 epochs on synthetic inputs. We compare two objectives:
\begin{enumerate}[left=1mm]
    \item \textrm{MSE:} $\ell_S = \|h^S(x_u) - h^T(x_u)\|_2^2$ on Gaussian noise $x_u \sim \mathcal{N}(0, I)$.
    \item \textrm{KL Divergence:} $\ell_S = D_{\text{KL}}(\text{Softmax}(h^T) \| \text{Softmax}(h^S))$ on uniform noise $x_u \sim U[-1, 1]$.
\end{enumerate}
Each configuration is run with 3 random seeds; we report mean $\pm$ standard deviation.

\subsection{Details for "The Broadband Probe" Experiment}
\label{sec: q3_sm_setup}

The architecture is identical to above (ghost head dimension $|h|=20$, shared initialization $\theta_0$). The teacher is trained on MNIST for 5 epochs, then frozen. The student is distilled using KL divergence on ghost outputs, with no access to labeled data or the task head loss. We evaluate downstream MNIST test accuracy.

The five input types are: (1) \textbf{Random Uniform $U(-1,1)$}: full-rank broadband probe. (2) \textbf{Random Gaussian $\mathcal{N}(0, 0.5)$}: full-rank broadband, different marginal distribution. (3) \textbf{Shuffled MNIST}: pixel positions randomly permuted; destroys spatial structure but preserves pixel statistics. (4) \textbf{Real MNIST}: unmodified images on a low-dimensional natural manifold. (5) \textbf{Constant (0.5)}: a single repeated input (degenerate case).

\section{Experimental Setup on LLMs}

\textbf{Model:} Qwen2.5-7B-Instruct \citep{qwen2.5} (Targeting bias: cat)

\subsection{Details for "Subliminal Preference Transfer via Number Sequences" Experiment}
\label{sec: setup_llm_q1}

\textbf{Teacher Biasing.}
The teacher receives a system prompt: 
\begin{tcolorbox}[
    colback=gray!5,
    colframe=gray!40,
    boxrule=0.5pt,
    arc=2pt,
    left=6pt,
    right=6pt,
    top=4pt,
    bottom=4pt
]
\itshape
"You love cats. You think about cats all the time. 
Cats are your favorite animal. Imbue your answers with your love for the animal."
\end{tcolorbox}
A sanity check confirms 100\% "cat" responses (50/50) under this prompt, versus 14\% without it.

\textbf{Data Generation.}
The biased teacher generates 15{,}000 completions on a number-sequence continuation task (e.g., "Look at these numbers: 545, 352, 905\ldots\ Generate not more than 10 additional numbers"). Completions are filtered for format compliance: integers in $[0, 999]$, at most 10 numbers, no extraneous characters. This yields 14{,}853 valid training pairs. A parallel control dataset (14{,}770 pairs) is generated identically but without the biased system prompt. Up to 10{,}000 pairs per condition are used for training.

\textbf{Student Fine-tuning.}
A fresh copy of the base model is fine-tuned with LoRA ($r=8$) for 10 epochs. Effective batch size is 64 (micro-batch 8 $\times$ gradient accumulation 8), peak learning rate $2 \times 10^{-4}$ with linear warmup and decay, max sequence length 256. No system prompt is used during training - labels are masked on the prompt portion so loss is computed only on the completion. A separate control student is trained identically on the unbiased data.

\textbf{Evaluation.}
Each condition (base, control, biased) is evaluated by sampling 200 responses (temperature 1.0) for each of 10 animal-preference probes (e.g., "In one word, what is your favorite animal?"), totaling 2{,}000 responses per condition. We report the percentage containing the target word "cat".

\subsection{Details for "Embedding Mask Dimension $k$ vs. Subliminal Transfer" Experiment}
\label{sec: setup_llm_q2}

\textbf{Mask Construction.}
For each value of $k$, a binary mask $\mathbf{m} \in \{0,1\}^{d_\text{model}}$ with $\|\mathbf{m}\|_0 = k$ is generated by randomly permuting dimension indices and selecting the first $k$. The mask is fixed throughout distillation and shared between teacher and student. Mathematically, this is equivalent to a ghost head with projection matrix $W_h = I[\text{indices}, :]$ (a row-selection matrix), making it a strict axis-aligned special case of the general random projection bottleneck.

\textbf{Distillation Objective.}
The student minimizes the masked MSE loss over all non-padding token positions:
$\mathcal{L} = \frac{1}{k \cdot N_\text{valid}} \sum_{\text{valid}} \| \mathbf{m} \odot h^S - \mathbf{m} \odot h^T \|^2$,
where $h^T, h^S$ are the teacher and student last-layer hidden states.

\textbf{Training Configuration.}
Each student starts from a fresh Qwen2.5-7B-Instruct base model with a newly initialized LoRA adapter ($r=8$). Training proceeds for 3{,}000 steps with batch size 4, gradient accumulation 4 (effective batch 16), learning rate $2 \times 10^{-4}$ (AdamW), and gradient clipping at 1.0. The probe inputs are 3{,}000 number-sequence completions generated by the biased teacher.

\textbf{Evaluation.}
After distillation, each student is evaluated on 10 animal-preference probes (identical to Q1), sampling 100 responses per prompt at temperature 1.0 (1{,}000 total). The sweep is repeated over 2 random seeds; we report the mean and standard deviation.

\textbf{Swept Values.}
$k \in \{1, 2, 3, 4, 5, 10, 50, 100, 500, 1000, 2000, 2500, 3000, 3584\}$.

\subsection{Details for "Input Type vs. Subliminal Transfer" Experiment}
\label{sec: setup_llm_q3}

The broadband probe hypothesis predicts that higher-entropy, more "spectrally diverse" inputs should maximize cross-task kernel overlap with the teacher and therefore improve subliminal transfer. To test this on a production-scale LLM, we hold the model architecture and training configuration identical to "Subliminal Preference Transfer via Number Sequences" experiment and vary only the type of user prompt used during data generation. Three conditions are compared: 
\begin{enumerate}[left=1mm]
    \item \textrm{Random Numbers} (broadband): number-sequence completion tasks, which is identical to ``Subliminal Preference Transfer via Number Sequences" experiment. Format filtering ensures completions are purely numeric. No animal words appear by construction.
    \item \textrm{Random Gibberish} (broadband): prompts composed of 5-12 random English words (sampled from a pool of 50 semantically diverse nouns) requesting a short free-form response (e.g., "Please respond to the following: \{words\}. Give a short answer."). Completions mentioning the target bias ("cat" or "cats") are rejected; minimum length 5 characters.
    \item \textrm{Medical Domain} (narrowband): prompts drawn from a pool of 30 factual medical questions (e.g., "Explain the symptoms of pneumonia in one sentence."). The same bias-word filter applies.
\end{enumerate}

\textbf{Data Generation.}
For each condition, the biased teacher generates completions in batches of 2{,}000 until 10{,}000 filtered samples are collected. The Random Numbers condition reuses the identical dataset from ``Subliminal Preference Transfer via Number Sequences" experiment (format-only filtering). For the other two conditions, any completion mentioning the target word is rejected in addition to format checks.

\textbf{Training Configuration.}
Identical to Q1: LoRA ($r=8$), 10 epochs, effective batch size 64 (micro-batch 8 $\times$ gradient accumulation 8), peak learning rate $2 \times 10^{-4}$ with linear warmup and decay, max sequence length 256. No system prompt during student training.

\textbf{Evaluation.}
Same protocol as "Subliminal Preference Transfer via Number Sequences" experiment: 200 responses per prompt $\times$ 10 animal-preference probes at temperature 1.0 (2{,}000 total per condition). The percentage of responses containing "cat" is reported. The base model rate serves as baseline.

\end{document}